\documentclass[letterpaper, 10 pt, journal, twoside]{IEEEtran}

\IEEEoverridecommandlockouts                              % This command is only
\usepackage{graphicx} % for pdf, bitmapped graphics files
\usepackage{amsmath} % assumes amsmath package installed
\usepackage{amssymb}  % assumes amsmath package installed
\usepackage{algorithm}
\usepackage{algorithmic}
\usepackage[dvipsnames]{xcolor}
\usepackage{multirow}
\usepackage{multicol}
\usepackage{tabularx}

\usepackage{FiraMono}
\makeatletter
\let\NAT@parse\undefined
\makeatother
\usepackage{float}
\graphicspath{{figures/}}
\usepackage{booktabs}

\usepackage[font=scriptsize,caption=false]{subfig}
\usepackage{hyperref}

\usepackage[export]{adjustbox}

\newcommand{\changes}[1]{{\color{black} {#1}}}

\newcommand{\pred}[1]{{\small\texttt{#1}}}
\newcommand{\var}[1]{{\textlangle\ignorespacesafterend #1\textrangle}}

\newcolumntype{P}[1]{>{\centering\arraybackslash}p{#1}}
\newcolumntype{M}[1]{>{\centering\arraybackslash}m{#1}}

\begin{document}

% \title{\LARGE \bf Long-Horizon Task and Motion Planning with \\ Functional Object-Oriented Networks}
% \title{\LARGE \bf Long-Horizon Manipulation Planning with \\ Functional Object-Oriented Networks}
\title{Long-Horizon Planning and Execution with Functional Object-Oriented Networks}

% \markboth{IEEE Robotics and Automation Letters. Preprint Version. Accepted June~2022}
% {Paulius \MakeLowercase{and} Agostini \MakeLowercase{\textit{et al.}}: Long-Horizon Planning and Execution with FOON} 

\author{
David Paulius$^{*}$, Alejandro Agostini$^{*}$, and Dongheui Lee\\
% <-this % stops a space\\
% \thanks{*This work was not supported by any organization}% <-this % stops a space
% \thanks{Manuscript received: April 1, 2023; Accepted: June 1, 2023.}
% \thanks{This paper was recommended for publication by Editor Aleksandra Faust upon evaluation of the Associate Editor and Reviewers' comments.
\thanks{
This research was funded by the Helmholtz Association and the Austrian Science Fund (FWF) Project M2659-N38. 
David is supported by the Office of Naval Research (ONR) under grant no. N00014-21-1-2584 and Echo Labs. 
This work began while the authors were members of the Human-centered Assistive Robotics group at the Technical University of Munich, Germany.
} 
\thanks{
David Paulius (Email:~\href{mailto:dpaulius@cs.brown.edu}{dpaulius@cs.brown.edu}) is affiliated with the Intelligent Robot Lab at Brown University, Rhode Island, United States.}
\thanks{Alejandro Agostini (Email:~\href{mailto:alejandro.agostini@uibk.ac.at}{alejandro.agostini@uibk.ac.at}) is affiliated with the Intelligent and Interactive Systems group at the Department of Computer Science, University of Innsbruck, Austria.}
\thanks{Dongheui Lee (Email:~\href{mailto:dongheui.lee@tuwien.ac.at}{dongheui.lee@tuwien.ac.at}) is affiliated with the Autonomous Systems group at TU Wien, Austria and also with the Institute of Robotics and Mechatronics, German Aerospace Center (DLR), Germany.}
\thanks{($^{*}${\it Alejandro Agostini and David Paulius are co-first authors.}) ({\it Corresponding authors: Alejandro Agostini and David Paulius.})}
% \thanks{Digital Object Identifier (DOI): see top of this page.}
}

\maketitle
\thispagestyle{empty}
\pagestyle{empty}

%%%%%%%%%%%%%%%%%%%%%%%%%%%%%%%%%%%%%%%%%%%%%%%%%%%%%%%%%%%%%%%%%%%%%%%%%%%%%%%%
\begin{abstract}
Following work on joint object-action representations, \textit{functional object-oriented networks} (FOON) were introduced as a knowledge graph representation for robots.
A FOON contains symbolic concepts useful to a robot's understanding of tasks and its environment for \textit{object-level planning}.
Prior to this work, little has been done to show how plans acquired from FOON can be executed by a robot, as the concepts in a FOON are too abstract for execution.
\changes{
We thereby introduce the idea of exploiting object-level knowledge as a FOON for task planning and execution.
Our approach automatically transforms FOON into PDDL and leverages off-the-shelf planners, action contexts, and robot skills in a hierarchical planning pipeline to generate executable task plans.}
We demonstrate our entire approach on long-horizon tasks in CoppeliaSim and show how learned action contexts can be extended to never-before-seen scenarios.

\end{abstract}
%%%%%%%%%%%%%%%%%%%%%%%%%%%%%%%%%%%%%%%%%%%%%%%%%%%%%%%%%%%%%%%%%%%%%%%%%%%%%%%%

\IEEEpeerreviewmaketitle

\vspace{-0.3cm}

\section{Introduction}

\IEEEPARstart{A}n ongoing trend in robotics research is the development of robots that can jointly understand human intention and action and execute manipulations for human domains.
A key component for such robots is a knowledge representation that allows a robot to understand its actions in a way that mirrors how humans communicate about action~\cite{paulius2019survey}.
Inspired by the theory of affordance~\cite{Gibson_1977} and prior work on joint object-action representation~\cite{SunRAS2013}, the \emph{functional object-oriented network} (FOON) was introduced as a knowledge graph representation for service robots~\cite{paulius2016functional,paulius2018functional}.
FOONs describe object-oriented manipulation actions through its nodes and edges and aims to be a high-level planning abstraction closer to human language and understanding.
% \todo{FOONs can be automatically generated from video demonstrations.}
They can be automatically created from video demonstrations~\cite{jelodar2018long}, and a set of FOONs can be merged into a single network from which knowledge can be quickly retrieved as plan sequences called task trees~\cite{paulius2016functional}.

% \todo{Why to keep using FOON instead of directly handcrafting planning operators?}
% This significantly simplifies the generation of knowledge representation for task planning compared to handcrafting planning operators one by one, as done traditionally. In this work we exploits this advantage by defining automatic mechanisms to transform FOON into PDDL notation for task planning and execution using a robotic platform. 
% \todo{What is the advantage of using PDDL POs over using only FOON?}
% Furthermore, the \textit{deconstruction} of FOON into planning operators allows generating sequences of functional units beyond the fixed sequences encoded in a FOON.

\begin{figure}[t]
    \centering
	\includegraphics[width=0.95\columnwidth]{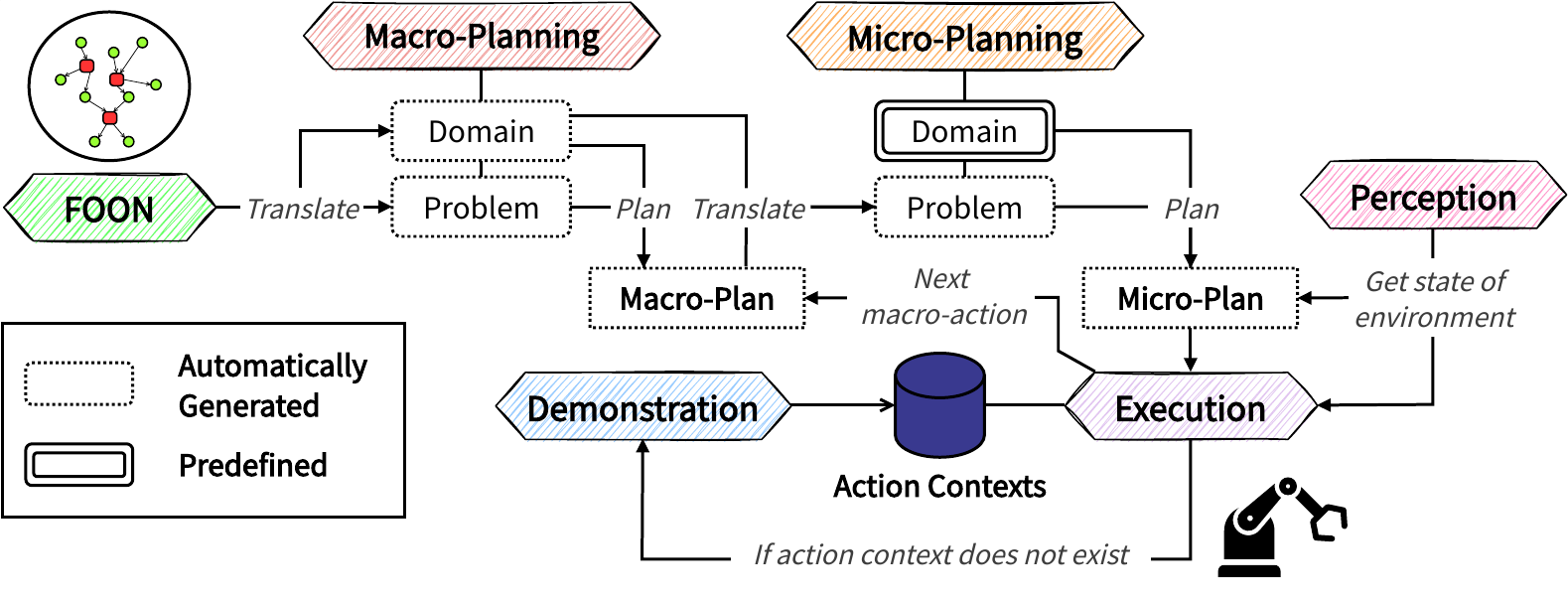}
 	\caption{Overview of our hierarchical planning approach. Domain-independent action sequences in FOON are automatically transformed into PDDL definitions, which are then used for deriving an object-level plan (\textit{macro-plan}) and task-level plan (\textit{micro-plan}). Each step in a macro-plan is grounded to the robot's action space and environment (via perception) as a micro-plan, which is a sequence of robot-executable skills based on action contexts~\cite{agostini2020manipulation}.\vspace{-0.5cm}
        }
	\label{fig:pipeline}
\end{figure}

\changes{
Although task plans extracted from FOON are too abstract for robot execution, FOON can significantly simplify the generation of knowledge representation for task and motion planning (TAMP) instead of handcrafting PDDL~\cite{mcdermott1998pddl} (short for \textit{Planning Domain Definition Language}), as done traditionally.
Indeed, a FOON is ideal for deriving \textit{object-level} plans that are agnostic to the robot and its environment, as opposed to \textit{task-level} plans, which considers robot and environment constraints~\cite{paulius2022objectlevel}.
Doing so requires grounding high-level semantic concepts in FOON to the low-level skills and parameters through which a robot interacts with or understands its actions and world~\cite{konidaris2019abstraction,kroemer2021review}.
For example, cooking recipes are object-level plans, but they require task-level plans to ground nouns to object instances in the world and verbs to robot skills.

Therefore, we introduce a two-level hierarchical task planning approach bootstrapped by FOON (Fig.~\ref{fig:pipeline}).
Our approach exploits object-level knowledge to automatically create PDDL planning definitions compatible with off-the-shelf planners and finds a sequence of executable skills, with which a robot can achieve object-level objectives.
Further, this approach \textit{deconstructs} a FOON into planning operators, allowing us to generate functional unit sequences beyond those fixed and encoded in a FOON.
Our contributions are as follows:}
%
% In this work, we exploit automatic mechanisms to transform FOON into PDDL.
% Furthermore, the \textit{deconstruction} of FOON into planning operators allows us to generate sequences of functional units beyond the fixed sequences encoded in a FOON.
%
% Therefore, to address task and motion planning (TAMP) using FOON, we introduce a hierarchical task planning approach to translate a FOON graph into a manipulation plan.
% Our algorithm creates a domain and problem definition in the PDDL planning language~\cite{mcdermott1998pddl} from a FOON, and they are used to find an appropriate sequence of low-level actions or primitives that can be executed by a robot to achieve the intended results of executing the graph from start to end~\cite{paulius2021roadmap}.
%
%
% \todo{
% \textbf{Points to address}:
% \begin{itemize}
%     \item We can automatically (???) map FOON object and state types to an object-centered representation. DONE.
% \end{itemize}
% }
%
%
%\subsection{Contributions}
%\noindent~
%
\begin{itemize}
    \item We introduce an approach to bootstrap task planning by automatically transforming a \textit{high-level}, symbolic FOON into \textit{low-level} planning problems in PDDL.

    \item We show how our approach finds plans for \textit{novel scenarios}, which may comprise random object configurations or ingredient sets, for the same high-level objective. 

    \item \changes{We show how our approach successfully executes long-horizon task plans by leveraging motion dependencies between actions and geometrical consistency via action contexts and an object-centered representation in PDDL.}
    % thanks to an object-centered PDDL representation that allows for the generation of geometrically consistent plans and to  that permits considering motion dependencies between plan steps for a smooth task execution.
    
    \item We show that our approach has a significantly lower time complexity over classical and HTN planning strategies.
\end{itemize}

% \begin{figure}[t]
% 	\centering
% 	\includegraphics[width=7cm]{fig-overview.pdf}
% 	\caption{Overview of the pipeline introduced in this work to translate a high-level FOON graph into low-level task planning problems in the PDDL planning language. Once decomposed, a low-level task plan can be executed using motion primitives associated with learned action contexts. \AG{Merge this fig with Fig. 3?}}
% 	\label{fig:overview}
% \end{figure}

% Furthermore, the FOON dataset is comprised of subgraphs whose contents were collected from manual annotation by human volunteers.
% To address these limitations, we propose a pipeline (shown as Fig.~\ref{fig:overview}) that integrates learning from demonstration (LfD) such that: 1) we can construct graphs directly from observation, and 2) we can integrate symbolic knowledge in FOON with physical aspects, such as a robot's motion primitives, object parameters, and trajectories.
% Although some work has been done to semi-automatically construct graphs from videos~\cite{jelodar2018long}, there is merit to learning in a teacher-student setting, where a human demonstrator can teach necessary skills to a robot to augment knowledge gathered from videos.

\section{Background}

\subsection{Functional Object-Oriented Networks (FOON)}
\label{sec:FOON}
Formally, a FOON is a bipartite graph $\mathcal{G} = \{\mathcal{O}, \mathcal{M}, \mathcal{E}\}$, where $\mathcal{O}$ and $\mathcal{M}$ refer to two types of nodes: \textit{object nodes} and \textit{motion nodes}.
Object nodes refer to objects used in tasks, including tools, utensils, ingredients or components, while motion nodes refer to actions that can be performed on said objects.
An object node $o \in \mathcal{O}$ is identified by its object type, its states, and, in some cases, its make-up of ingredients or components; a motion node $m \in \mathcal{M}$ is identified by an action type, which can refer to a manipulation (e.g., pouring, cutting, or mixing) or non-manipulation action (e.g., frying or baking).
Edges ($e \in \mathcal{E}$) connect these nodes to one another.

\begin{figure}[t]
    \centering
    \includegraphics[width=7cm]{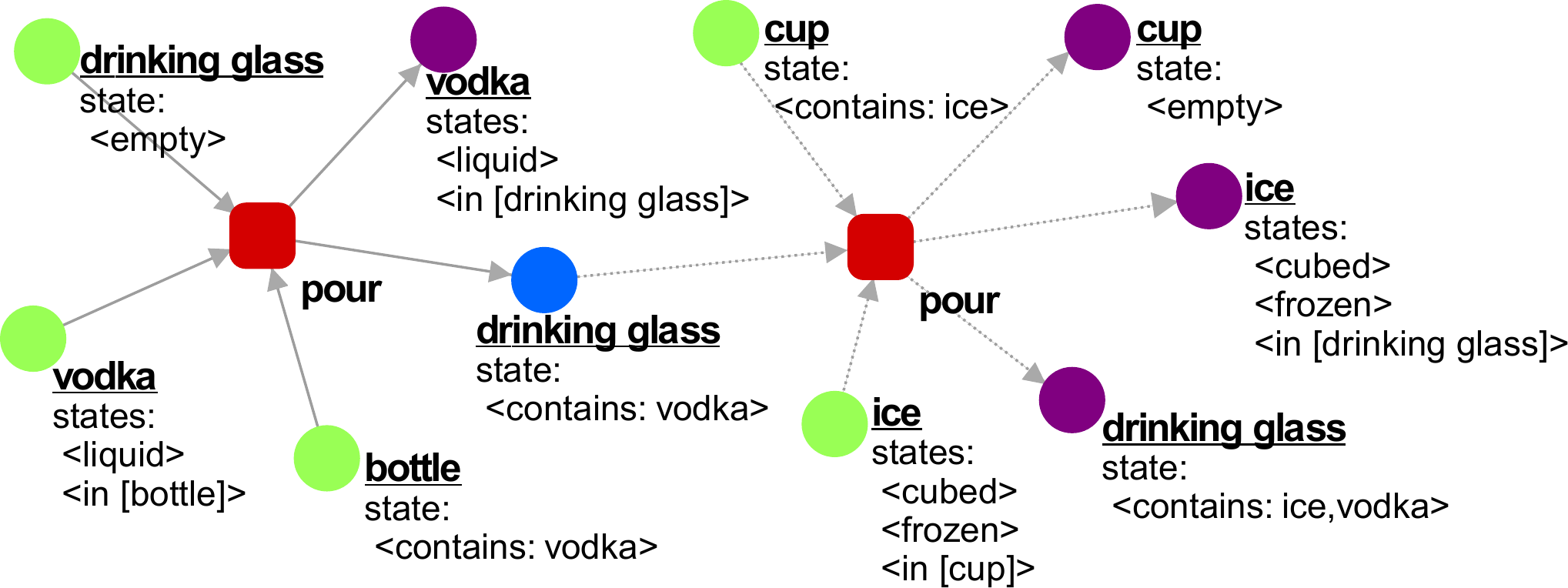}
    \caption{
        Illustration of two functional units for pouring vodka and ice into a glass (best viewed in colour).     
        Each functional unit is discernible by edge style (solid and dashed respectively).
        Object and motion nodes are denoted by circles and squares respectively.
        Input-only and output-only nodes are shown in green and purple respectively, while nodes that are both are shown in blue.
        \vspace{-0.6cm}
    }
    \label{fig:unit}
\end{figure}

Objects may take on new states as a result of executing actions. 
State transitions are conveyed through edges ($\mathcal{E}$) to form \textit{functional units} (denoted as $\mathcal{FU}$), which describe object nodes before and after an action takes place.
Specifically, a functional unit $\mathcal{FU} = \{\mathcal{O}_{in}, \mathcal{O}_{out}, m\}$ contains a set of input nodes $\mathcal{O}_{in}$, a set of output nodes $\mathcal{O}_{out}$, and an intermediary action node $m$, comparable to the \textit{precondition-action-effect} structure of planning operators (POs) in classical planning~\cite{ghallab_nau_traverso_2016}.
% Functional units can be linked to planning operators (PO) that are used in planning languages such as PDDL~\cite{mcdermott1998pddl}.
A robot can use a FOON to identify states that determine when an action is completed.
Fig. \ref{fig:unit} shows two functional units describing a sequence of pouring vodka and ice into a drinking glass. 
There are notably several object types with multiple node instances, as these object states will change as a result of execution. 
%
% ALE deleted: This is analogous to Petri Nets~\cite{Petri:2008}, where the firing of transitions cause a change in input place nodes.
%
Each functional unit has the same motion node label of \textit{pour}, yet the objects and effects of each action differ, thus treating them as two separate actions.

% \subsubsection{Creating a FOON}
FOONs are created by annotating action from observation, such as video demonstrations.
% , or, as we plan to explore as future work, demonstrations from a human teacher.
We note the objects, actions, and state changes required to achieve a specific goal, such as a recipe, during annotation.
This results in a \textit{subgraph}, which is a sequence of functional units (and their respective objects and actions) to fulfill the given goal.
Two or more subgraphs can be merged to form a \textit{universal} FOON.
% This merging procedure consolidates all instances of object nodes and removes duplicate functional units that are common across different subgraphs~\cite{paulius2016functional}.
Presently, the FOON dataset provides 140 subgraph {annotations of recipes} 
% \AG{For which type of task? Recipies?} 
with which a universal FOON can be created; these annotations along with helper code are publicly available for use.\footnote{FOON API and Dataset -- \url{https://github.com/davidpaulius/foon_api}}
\vspace{-0.3cm}
%
% \rII
\subsection{Task Planning}
\label{sec:taskplanning}
We adopt the traditional approach for task planning~\cite{ghallab_nau_traverso_2016} by defining a set of objects (e.g., \pred{cup} or \pred{bowl}) and predicates that encode object properties or relations (e.g., \pred{(on table cup)} -- the cup is on the table).
Each predicate can be true or false depending on whether these attributes are observed in the scene. 
The \textit{symbolic} state $\mathcal{S}$ is defined by a set of predicates describing the object configuration in a scenario. 
Planning operators (POs) describe the changes in the symbolic state via actions and are encoded in the traditional \textit{precondition-action-effect} notation using PDDL~\cite{mcdermott1998pddl}. Preconditions comprise the predicates that change by the execution of the PO as well as those that are necessary for these changes to occur. Effects, in turn, describe the changes in the symbolic state after the PO execution as predicates. 
Fig.~\ref{fig:micro-PO} provides examples of POs written in PDDL notation.
The name of a PO is a {\it symbolic action} and may contain arguments to ground the predicates to the preconditions and effects. 
In task planning, a planner uses a description of the \textit{initial state} ($s \in \mathcal{S}$) and a \textit{goal} definition ($g$) as a set of grounded predicates that should be observed after execution. 
The planner carries out a heuristic search with these elements by generating causal graphs from the preconditions and effects of POs and yields a sequence of actions called a {\it plan} that produces changes in $s$ necessary to obtain $g$. 

% \begin{figure*}[t]
%     \vspace{4px}
%     \centering
% 	\includegraphics[width=12cm,clip,trim={0.5cm 0.2cm 0 0cm}]{fig-pipeline-3.pdf}
%  	\caption{Illustration of the entire procedure for: 1) translating a FOON graph into its macro-level domain and problem representation as PDDL, 2) decomposing each macro-level action (i.e., planning operator) into its respective micro-level PDDL problem, 3) planning at the micro-level using pre-defined motor skills (defined in the micro-level domain), and 4) executing the acquired micro-level task plan using action contexts learned from demonstration.
%  	}
% 	\label{fig:pipeline}
% \end{figure*}
%
\vspace{-0.2cm}
\subsection{Related Work}
There are many notable works that aim to represent knowledge for robots in a way that encourages abstraction for task and motion planning.
Frameworks such as {\sc{KnowRob}}~\cite{tenorth2017representations} combine knowledge bases with a query processing engine to allow reasoning over beliefs of the world.
Tenorth et al. showed how a robot can use {\sc{KnowRob}} to prepare meals, such as pancakes, and form queries over object or action properties.
However, they focused on structurally defining this knowledge base and inferring object locations rather than storing or retrieving recipes or task sequences as possible with FOON.
Rather, FOON is best used as a schema with reasoning engines or knowledge bases like {\sc{KnowRob}}.
% , which are tied to low-level robotic properties.
Ramirez-Amaro et al.~\cite{ramirez2017transferring} investigated how semantic knowledge can be learned from demonstration and then used in planning to imitate demonstrated tasks, such as making pancakes and a sandwich.
Although our work does not adopt the same degree of object and activity recognition,  knowledge in FOON is agnostic to the robot, and it is only through planning that we obtain a robot-specific task plan suited to its present environment state.

Kaelbling and Lozano-P\'{e}rez interleave hierarchical planning with execution using highly abstract behaviours for task planning to accelerate plan generation, but at the expense of planning impasses at execution time~\cite{kaelbling2011hierarchical}. 
Our approach leverages relevant geometrical constraints at the task planning level to exploit the computational efficiency of planners in generating feasible manipulation plans.
Logic programming task planners search for solutions directly in the plan space, rather than in the state space as classic planners, to generate feasible task plans using geometrical constraints~\cite{toussaint2015logic}. 
However, these approaches require computationally demanding optimization processes on whole plans using complex dynamic models, making them less suitable for solving long-horizon optimization problems.
% 
% Our approach, instead, does not require the robot dynamics and is able to generate plans at low computational costs using off-the-shelf linear planners. Although we consider motion dependencies between consecutive actions rather than in the entire plan, as in logic programming, it provides an appealing low-complexity alternative to these methods.
%
Other approaches use semantic descriptions of geometrical constraints to evaluate motion feasibility of single actions~\cite{dantam2018incremental} or sequences of actions~\cite{garrett2020pddlstream} that are assessed during task planning using conventional state-based planners. A task planner finds candidate plans based on these constraints, while a sampling-based motion planner checks action feasibility using geometric reasoning. 
% Garrett et al. proposed a new PDDL domain, the PDDLstream~ \cite{garrett2020pddlstream}, which includes sequences of symbolic constraints (streams) that are assessed during task planning using conventional state-based planners. PDDLStream requires handcrafting several ad-hoc functions that map object parameters to arbitrary predicates in a ``black-box" approach. 
%
Instead, we use object-centered predicates to propagate geometrical constraints during task planning in terms of standard relational names that easily map to object poses without the need of external heuristics for geometric reasoning.

Hierarchical task networks (HTN)~\cite{ghallab_nau_traverso_2016} share many similarities with our approach.
HTNs represent abstract tasks called \textit{methods}, comprising a sequence of sub-tasks that are executed in a reactive manner.
A planning problem is defined as solving a series of tasks.
Such tasks are akin to FOON functional units, as they require a sequence of lower-level actions to accomplish a task.
However, methods must be defined for all possible ways of executing tasks, which is not practical in real-world settings.
Our FOON-based approach treats each high-level task as its own planning problem, which greatly reduces the planning time complexity without defining fixed-ordered methods.
% Using schematic knowledge from FOON allows us to creating planning definitions based on the state of the robot's workspace. 

% Similarly, previous work investigated how to encode macro and primitive planning operators into a single linear planning domain for the execution of robotic tasks~\cite{agostini2008action}.
Previous work explored the encoding of {\it macro} planning operators into primitive operators for the execution of robotic tasks, combining macro operators and primitives into a single linear planning domain~\cite{agostini2008action} or combining linear planning with reinforcement learning for executing primitives~\cite{quack2015simultaneously}. However, as with HTNs, macro operators are associated with a fixed sequence of primitive operators that are executed in a {reactive} manner.
Manipulation action trees by Yang et al.~\cite{yang2014manipulation} represent robotic manipulation as a tree for planning and execution.
Zhang and Nikolaidis~\cite{zhang2019robot} proposed executable task graphs, which describe what the robot must do to replicate actions observed from cooking videos, for multi-robot collaboration. However, as their focus was on imitating behaviours from demonstration, they do not show how these graphs could be adapted to novel scenarios as possible with our approach.
% However, the aforementioned task graphs are very domain-specific and do not show the same degree of flexibility, where they can be used in symbolic or object-level planning across various scenarios, as possible with FOON.

% This paper is organized as follows: in Sec.~\ref{sec:FOON}, we give a short background on the FOON structure, terminology, and algorithms for the retrieval of a FOON task tree.
% In Sec.~\ref{sec:tamp}, we introduce our approach to translate a FOON graph into PDDL for hierarchical task planning.
% In Sec.~\ref{sec:exe}, we then discuss how manipulation plans can be executed based on primitives learned from demonstration.
% In Sec.~\ref{sec:exp}, we evaluate our approach via experiments in simulation to show how a FOON graph can be adapted to new scenarios.
% Finally, in Sec. \ref{sec:con}, we summarize our ideas and provide insight to future work.

\vspace{-0.2cm}
\section{Task Planning with FOON}
\label{sec:tamp}
\changes{
% Up to this point, using FOONs to bootstrap robotic execution has not been investigated.
Object-level representations like FOON are ideal for bootstrapping TAMP, as they can be applied across robots and domains.
% \todo{FOON permits a direct transformation to PDDL notation for task planning as it is grounded in precondition and effects articulated by functional units.}
%This transformation is done to take advantage of existing solvers/planners to ground FOON into executable tasks for service robotics. DONE.
Further, functional units integrate well into PDDL notation, as they are encoded using a precondition-action-effect notation that can be directly transformed into PDDL predicates and planning operators. 
% compatible with planning operators of classic planning as predicates, preconditions, and effects can be extracted from FOON.
However, we must ground the \textit{domain-independent}, object-level knowledge to how the robot views or interacts with the world as a \textit{domain-specific} representation, where concepts in FOON are grounded to the physical world, relevant object properties, and robot actions.}

To use object-level knowledge in FOON for task planning, we devise a two-level hierarchical planning approach (Fig.~\ref{fig:pipeline}). 
At the top, {\it macro}-planning finds a plan skeleton (\textit{macro-plan} -- $\mathcal{P}_{M}$) to prepare a recipe. 
At the bottom, {\it micro}-planning finds a sequence of robot skills (\textit{micro-plan} -- $\mathcal{P}_{\mu}$) to execute each FOON action (i.e., macro-plan step) in a given scenario and fulfill the high-level objectives of a macro-plan.
% We illustrate our approach as .
% }

% However, we would like to point out that the FOON representation can be translated into PDDL notation using one-to-one transformations of object relations and properties, without information loss, as specified in Sec. III.  In this light, we believe that the optimality and efficiency of our two-level planning approach depends, on the one hand, on the macro-planning approach with FOON for the elaboration of recipes using objects and ingredients present in the scenario (object-level planning), and, on the other hand, on the manipulation planning method using an object-centric representation (task-level planning). The performance of these two planning approaches have been demonstrated individually in the previous contributions \cite{paulius2016functional} and \cite{agostini2020manipulation}, respectively.

% \rII
\begin{figure}[t]
    \centering
    \subfloat[Functional unit for pouring vodka]{
     \adjustbox{valign=c}{\includegraphics[width=4.19cm]{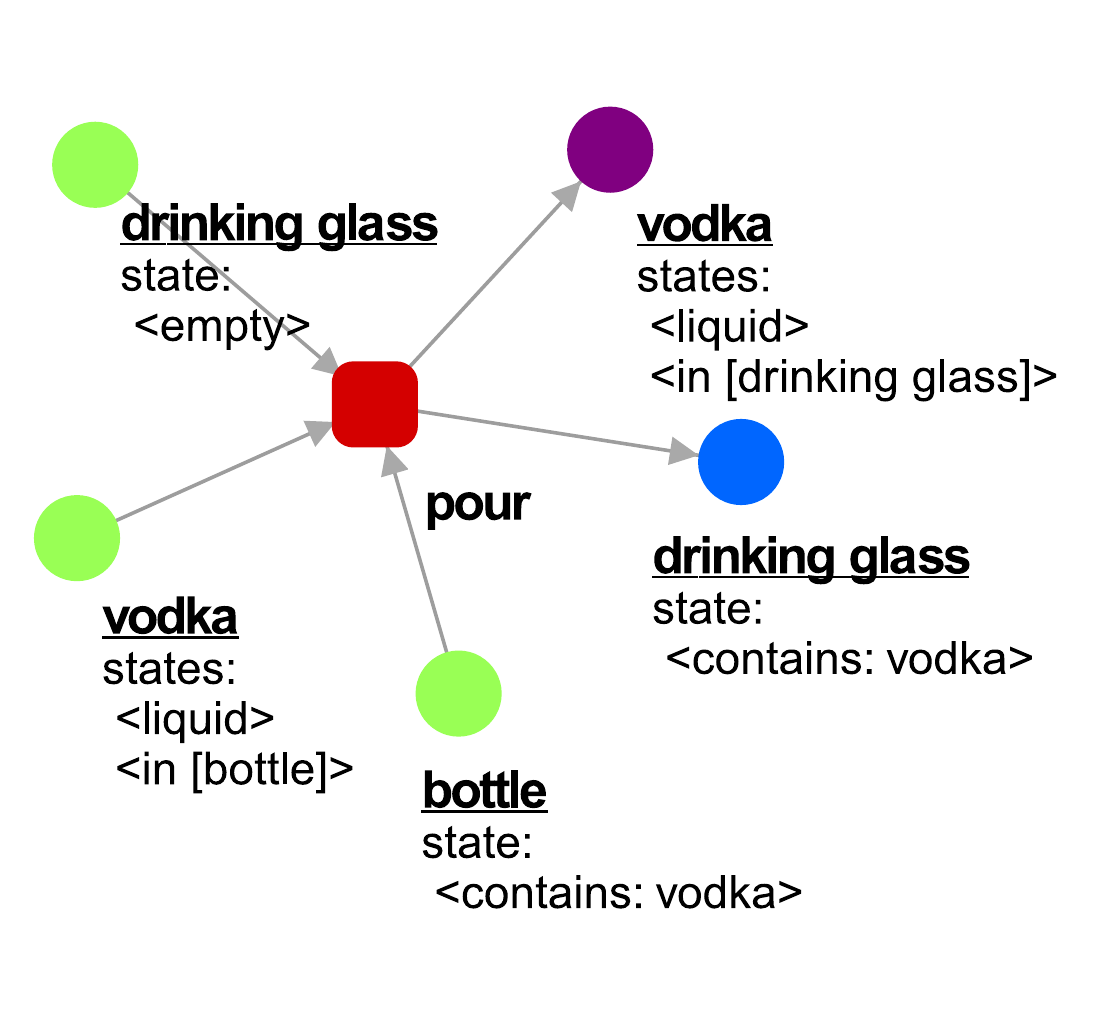}}
     \label{fig:translation-a}}
    \subfloat[Equivalent Macro-Level Definition]{
    \texttt{\tiny
    \adjustbox{valign=c}{\begin{tabular}{l}
    (\textbf{:action} pour\_vodka\_0 \\
	\indent~\textbf{:parameters} () \\
	\indent~\textbf{:precondition} (and \\
	\indent~\indent~(under bottle table) \\
	\indent~\indent~(on table bottle) \\
	\indent~\indent~(in bottle vodka) \\
	\indent~\indent~(under vodka bottle) \\
	\indent~\indent~(in drinking\_glass air) \\
	\indent~\indent~(under drinking\_glass table) \\
	\indent~\indent~(on table drinking\_glass) ) \\
	\indent~\textbf{:effect} (and \\
    \indent~\indent~(in drinking\_glass vodka) \\
	\indent~\indent~(under vodka drinking\_glass) \\
	\indent~\indent~(under bottle table) \\
	\indent~\indent~(in bottle vodka) \\
	\indent~\indent~(under drinking\_glass table) \\
	\indent~\indent~(on table drinking\_glass) \\
	\indent~\indent~(not (in drinking\_glass air)) ) )
    \end{tabular}
    }
    }\label{fig:translation-b}}
    \caption{Example of a functional unit for \textit{pouring} vodka into a drinking glass and its equivalent macro-PO in PDDL. 
    Note here that vodka remains in the bottle because the bottle is not emptied as in the functional unit definition.
    \vspace{-0.3cm}
    }
    \label{fig:translation}
\end{figure}

%\changes{
\vspace{-0.3cm}
\subsection{Macro-level Planning}
\changes{
\subsubsection{Overview}
The aim of macro-planning is to find a plan schema with which we can perform micro-planning; this is equivalent to finding an \textit{object-level plan} that describes how to use objects to solve a high-level objective.
In prior work, we defined a search algorithm that combines breadth-first and depth-first search approaches to directly find a task tree from a universal FOON~\cite{paulius2016functional}. 
However, in this work, we adopt an alternative strategy that transforms FOON into PDDL (example shown in Fig.~\ref{fig:translation}) and then derives solutions from a linear planner to generate a \textit{macro-plan}.
%
% \todo{Automatic generation of PDDL from FOON.}
We encode each functional unit as planning operators in PDDL, thus \textit{deconstructing} FOON into individual units from which we generate sequences beyond those fixed in a FOON.

% Functional units of FOON are already encoded using a precondition-action-effect notation that can be directly mapped into planning operators using a PDDL notation. 
% \todo{What is the advantage of using PDDL POs over using only FOON?}
% The translation from FOON to PDDL notation as well as the sequence generation is done automatically as explained below.
% However, similar to a task tree~\cite{paulius2016functional}, a macro-plan is still not directly executable by a robot.
}

\changes{
\subsubsection{Macro-level Translation}
% \todo{Automatic generation of PDDL from FOON.}
To perform macro-planning, we require domain and problem definitions compatible with linear planners.
We do this by \textit{automatically} parsing a FOON: object nodes define object names and predicates, while functional units define macro-planning operators (\textit{macro-PO}).
An object node $o \in \mathcal{O}$ is defined by its type and state attributes (e.g., a \textit{drinking glass} (type) is \textit{empty} (attribute) -- see Fig.~\ref{fig:unit}).
An object $o$ is characterized by one or more predicates of the following two types: 1) \textit{object-centered predicates}~\cite{agostini2020manipulation,agostini2020efficient}, which describe geometrical relations or properties for characterizing object configuration spaces for motion planning (e.g., defining what is \textit{in} or \textit{on} objects for manipulation); and 2) \textit{state of matter} predicates, which describe an object's physical state, which is useful for identifying a high-level objective (e.g., an object changes from \textit{whole} to \textit{chopped}).}
%
% To transform FOON into a PDDL representation suitable for recipe execution, we need to define, on the one hand, geometrical predicates that allow for a consistent propagation of geometric changes with actions during plan generation and execution.
%
% To this end, we use the same approach from prior work~\cite{agostini2020manipulation,agostini2020efficient} that characterize the object configuration space for manipulation planning using \textit{object-centered predicates}. 

% \todo{Geometric completeness.}
Object-centered predicates describe poses or locations of objects from each object's perspective as they relate to other objects within the robot's environment, allowing us to consistently represent and propagate geometrical constraints during the heuristic search, thus rendering geometrically feasible plans.
These predicates have the form of {\pred{(\var{rel} \var{obj\_1} \var{obj\_2})}} (see Fig. \ref{fig:translation-b}), where {\pred{\var{rel}}} refers to the spatial relation type, while {\pred{\var{obj\_1}}} and {\pred{\var{obj\_2}}} refer to the focal object and relative object respectively.
We use the spatial relations \textit{in}, \textit{on}, and \textit{under}, as these are typically attributed to object nodes in FOON (see Fig.~\ref{fig:translation-a}). 
For example, we define predicates such as {\pred{(on table \var{obj})}} and {\pred{(under \var{obj} table)}} to characterize the geometric changes taking place with pick-and-place actions of objects on the table.\footnote{
These object-table relations indicate that the object is \pred{on} the table and that the table is {\tiny \pred{under}} the object to consistently map table-object relations from the effects of a macro-PO to a goal for micro-planning (see Sec. \ref{sec:micro-planning}).} 
Additionally, we adopt the convention from prior work~\cite{agostini2020manipulation} to describe an \textit{empty} object as it containing {\pred{air}} (i.e., {\pred{(in \var{obj} air)}}).
% By default, if an object node has no states of this type \AG{which "type" do you refer to?}, we assume that it is simply present on the working surface (viz., \pred{table}).
\changes{By default, if an object node has no \textit{on} state relations, we assume that it is simply present on the working surface (viz., \pred{table}).
%
% In the case where an object \textit{contains} other objects, this is explicitly translated using the \pred{in} and \pred{under} relations (e.g., \pred{(in bottle vodka)} and \pred{(under vodka bottle)} for a bottle of vodka in Fig.~\ref{fig:translation}) \AG{Is under vodka bottle really necessary? We are assuming here that a liquid has a pose..., the Rs may complain.}.
In the case where an object \textit{contains} other objects, this is explicitly translated using the \pred{in} and \pred{under} relations (e.g., \pred{(in bottle vodka)} and \pred{(under vodka bottle)} for a bottle of vodka in Fig.~\ref{fig:translation}.}
%
% }
State of matter predicates characterize the physical properties of objects that are temporally relevant for cooking.
For instance, a \textit{whole} object becomes \textit{sliced} as a recipe progresses.
Several states in FOON have been identified in related work on state recognition for cooking~\cite{jelodar2019joint}. 
% Predicates for such states take the form of {\pred{(\var{rel} \var{obj})}}, where {\pred{\var{rel}}} refers to the state type and {\pred{\var{obj}}} refers to the focal object \AG{The argument name "rel" is misleading. It seems to be for object relation (rel) but it is actually for object state. Maybe \pred{\var{status}} or something like this?}. 
Predicates for such states take the form of {\pred{(\var{som} \var{obj})}}, where {\pred{\var{som}}} refers to the state of matter type and {\pred{\var{obj}}} refers to the focal object. 
\changes{Examples of these states and their respective predicates are {\pred{is-whole}} for the \textit{whole} state, {\pred{is-sliced}} for the \textit{sliced} state, and {\pred{is-mixed}} for the \textit{mixed} state.}
% \AG{The following sentence was not marked as changed and it was added now to address one of the most critical concerns of Rs and AE: automatic transformation and completeness. It also seems that other changes have been unmarked in the rest of the paper...could it be? Any reason behind this?}
\changes{
% \todo{Automatic transformation FOON-PDDL and completeness.}
The translation from FOON to macro-level PDDL notation is automatic and complete, as spatial relations and physical states are unambiguously mapped to object-centered and state of matter predicates respectively, provided that all states compatible with and related to special skills (e.g., chopping, slicing, mixing) are defined.
%
% It is also possible that certain states in FOON are disregarded in PDDL; for instance, the execution
}

\subsubsection{Finding a macro-plan}
We create macro-level domain and problem definitions for planning using the aforementioned predicate types.
To construct a macro-domain definition, we transform each functional unit $\mathcal{FU}$ into macro-planning operators by translating the objects in $\{\mathcal{O}_{in}, \mathcal{O}_{out}\} \in \mathcal{FU}$ into precondition and effect predicates.
Each macro-PO is assigned a name given by the motion node $m \in \mathcal{FU}$.
% $\{\mathcal{O}_{in}, \mathcal{O}_{out} \}$ $\forall$ $\mathcal{FU}$ into {\it macro} planning operators (\textit{macro-PO}) with name given by the $\mathcal{FU}$ name $n$, 
% Each macro-PO is then used to construct a 
\changes{
To construct a macro-problem definition, we define the initial state ($s_{M}$) as predicates describing objects initially available for use (i.e., no incoming edges), while we define the macro-goal ($g_{M}$) as predicates describing the desired final state from the goal node (e.g., \{\pred{(in drinking\_glass ice)}, \pred{(in drinking\_glass vodka)}\} $\in g_{M}$ as in Fig.~\ref{fig:unit}).
After the domain (macro-POs) and problem  (initial state and goal)
% ($s^{mac}_{ini}$ and $g^{mac}$) 
are defined, we can use a linear planner to find a macro-plan $\mathcal{P}_{M}$, which contains a sequence of steps (functional units) that should be fulfilled when preparing a recipe.
%The acquired macro-plan captures a
However, a micro-plan is required to execute each action $A \in \mathcal{P}_{M}$, as each action $A$ has yet to be grounded to a robot's action set. 
% \todo{Completeness.}
Transforming a FOON into PDDL macro-definitions is achieved without information loss, thus
% , only adding new predicates that confirm the availability of objects in the scenario for recipe preparation.
% To put it differently, each object can be traced to its related predicates in PDDL, as each functional unit is defined as originally seen in FOON.
% This preserves the completeness of the original FOON and guarantees finding identical solutions when using the same heuristics in both representations~\cite{paulius2016functional}.
preserving the completeness~\cite{ghallab_nau_traverso_2016} of the original FOON and guarantees finding identical solutions, granted that all required objects for a recipe are present~\cite{paulius2016functional}.
}

\vspace{-0.2cm}
\subsection{Micro-level Planning}
\label{sec:micro-planning}
\changes{
\subsubsection{Overview}
Once a macro-plan ($\mathcal{P}_{M}$) has been found, we must perform \textit{micro-planning} to generate a micro-plan ($\mathcal{P}_{\mu}$) that is executable by a robot.
In this process, we treat each functional unit (i.e., $A \in \mathcal{P}_{M}$) as a micro-planning problem, whose initial state ($s_{\mu}$) and goal ($g_{\mu}$) are taken directly from a macro-PO and grounded to objects and states via perception.
Along with the problem definition, we must define a micro-domain that details all robot-executable primitives, their necessary preconditions, and their resulting effects as micro-planning operators (\textit{micro-PO} -- see Fig.~\ref{fig:micro-PO}).
Therefore, with both micro-planning domain and problem definitions, we can acquire a manipulation plan that breaks down each macro-plan action into a sequence of realizable actions.}
% For instance, a micro-plan for a functional unit for \textit{pouring} in FOON (such as Fig.~\ref{fig:translation-a}) could be found that consists of a simple action sequence (\textit{micro-PO}): \textit{pick} a source container, \textit{pour} from source to target container, and \textit{place} the source container to free the robot's gripper.

\begin{figure}[t]
    \centering
    \subfloat[Pick]{
    \texttt{\tiny
    \begin{tabular}[t]{l}
      	(\textbf{:action} pick \\
    	\indent~\textbf{:parameters} ( \\
        \indent~\indent~\textit{?obj} - object \\
        \indent~\indent~\textit{?surface} - object ) \\
    	\indent~\textbf{:precondition} (and \\
    	\indent~\indent~(on \textit{?obj} air) \\
    	\indent~\indent~(under \textit{?obj} \textit{?surface}) \\
    	\indent~\indent~(on \textit{?surface} \textit{?obj}) \\
    	\indent~\indent~(in hand air) ) \\
    	\indent~\textbf{:effect} (and \\
    	\indent~\indent~(on \textit{?obj} hand) \\
    	\indent~\indent~(in \textit{?hand} \textit{?obj}) \\
    	\indent~\indent~(under \textit{?obj} air) \\
    	\indent~\indent~(on \textit{?surface} air) \\
    	\indent~\indent~(not (in \textit{?hand} air)) \\
    	\indent~\indent~(not (on \textit{?obj} air)) \\
    	\indent~\indent~(not (under \textit{?obj} \textit{?surface})) \\
    	\indent~\indent~(not (on \textit{?surface} \textit{?obj})) ) )\\
    	% )\\
    \end{tabular}
    }
    }
    \subfloat[Place]{
    \texttt{\tiny
    \begin{tabular}[t]{l}
      	(\textbf{:action} place \\
    	\indent~\textbf{:parameters} ( \\
        \indent~\indent~\textit{?obj} - object \\
        \indent~\indent~\textit{?surface} - object ) \\
    	\indent~\textbf{:precondition} (and \\
    	\indent~\indent~(on \textit{?obj} hand) \\
    	\indent~\indent~(under \textit{?obj} air) \\
    	\indent~\indent~(on \textit{?surface} air) \\
    	\indent~\indent~(in hand \textit{?obj}) \\
    	\indent~\textbf{:effect} (and \\
    	\indent~\indent~(on \textit{?obj} air) \\
    	\indent~\indent~(in \textit{?hand} air) \\
    	\indent~\indent~(under \textit{?obj} \textit{?surface}) \\
    	\indent~\indent~(on \textit{?surface} \textit{?obj}) \\
    	\indent~\indent~(not (in \textit{?hand} \textit{?obj})) \\
    	\indent~\indent~(not (on \textit{?obj} hand)) \\
    	\indent~\indent~(not (under \textit{?obj} air)) \\
    	\indent~\indent~(not (on \textit{?surface} air)) ) )\\
    	% )\\
    \end{tabular}
    }
    }
    \caption{Examples of \textit{micro-PO} action definitions in PDDL notation defined using object-centered predicates \cite{agostini2020efficient}.
    To account for object sizes (Sec.~\ref{sec:exe}), we defined various \textit{place} POs for small, long, and wide objects.\vspace{-0.25cm}
    }
    \label{fig:micro-PO}
\end{figure}

% We also provide an example of how a \textit{macro-PO} in a macro-level plan can be broken down into \textit{micro-PO} steps as Fig.~\ref{fig:macro_to_micro}.
% \rII
% {
% To generate micro-plans, we use the same approach as before from prior work~\cite{agostini2020manipulation,agostini2020efficient} that characterize the object configuration space for manipulation planning using \textit{object-centered predicates}. Object-centered predicates are used to describe poses or locations of objects from each object's perspective as they relate to other objects within the robot's environment, allowing us to consistently represent and propagate geometrical constraints during the heuristic search, thus rendering geometrically feasible plans.}
% These predicates have the form of {\pred{(\var{rel} \var{obj\_1} \var{obj\_2})}}, where {\pred{\var{rel}}} refers to the spatial relation type, while {\pred{\var{obj\_1}}} and {\pred{\var{obj\_2}}} refers to the focal object and relative object respectively.
% We use the relations \textit{in}, \textit{on}, and \textit{under}, as these are typically attributed to FOON object nodes.
% For instance, the predicate {\pred{(in bowl tomato)}} means that a \textit{tomato} is inside of a \textit{bowl}.
% Additionally, we adopt the convention from prior work~\cite{agostini2020manipulation} to describe an \textit{empty} object as it containing {\pred{air}} (i.e., {\pred{(in \var{obj} air)}}).
%

\changes{
\subsubsection{Micro-level Translation}
% We need micro-level domain and problem definitions in order to find a micro-plan.
We create a micro-problem corresponding to a macro-PO definition for each step of a macro-plan (i.e., $A \in \mathcal{P}_{M}$): each precondition and effect predicate forms the initial state ($s_{\mu}$) and goal ($g_{\mu}$) predicates.
As with macro-definitions, we use object-centered predicates and state of matter predicates to characterize the object configuration space for manipulation and physical states that objects undergo as a result of execution, respectively. 
However, these predicates must be grounded to how the robot perceives or interacts with its world. 
%
% \todo{Completeness.}
True and false values of object-centered predicates are directly obtained from object 3D poses and bounding boxes to provide a geometrically complete description of the object configuration space, which allows for a consistent propagation of geometric changes with actions for the generation of feasible plans \cite{agostini2020manipulation}. 
% \todo{No handcraft of physical constraints.}
Note that the perception mechanisms automatically generate predicates that describe the physical constraints for execution, such as \pred{(on table \var{obj})} and \pred{(in bowl air)}. 
% At the micro-level, 
We make some assumptions to guarantee unambiguous grounding and completeness.
First, we assume that objects are graspable by the robot from free surfaces.
We represent a robot's end-effector with a micro-level object \pred{hand}, which can be empty or not (i.e., \pred{(in hand air)} or \pred{(in hand \var{obj})}).
% and \pred{(on \var{obj} hand)}).
A robot can pick up an object from its top if no obstacles are on top of it (i.e., \pred{(on \var{obj} air)}, which transforms to \pred{(on \var{obj} hand)} after picking).  
We also treat the working surface (i.e., \pred{table}) as a grid of tiles, upon which objects may be present.
% , to ensure geometrical feasibility in pick-and-place actions.
These cells are present in our experiments (Sec.~\ref{sec:exp}), where we use tiles of varying sizes for different objects.
% Grounding is further done with functions that map object parameters (3D poses and bounding boxes) to true and false values of object-centered predicates as in prior work~\cite{agostini2020manipulation}. 
Further, macro-level object names are grounded to micro-level object instances, which are observable and usable by the robot (e.g., in the macro-level predicate \pred{(in bottle vodka)}, a \pred{bottle} object maps to a \pred{bottle\_vodka} instance).
These mechanisms guarantee that we find geometrically feasible and complete micro-plans.}

A micro-domain contains micro-POs that capture physical preconditions and expected effects of executable skills (e.g., pick, place, pour) in terms of object-centered and state relations.
These micro-POs consider constraints like the robot's hand (empty or not empty) and other aspects such as the position and orientation of objects and the available surfaces for robot-object and object-object interactions through the virtual object \pred{air}.
For this work, we manually define micro-level actions; examples of {micro-PO} are shown in Fig.~\ref{fig:micro-PO}, and further examples can be found in previous work~\cite{agostini2020manipulation}.

\changes{
\subsubsection{Finding a micro-plan}
With a macro-plan, we can acquire a micro-plan, which comprises micro-PO sequences for each macro-PO action, using an off-the-shelf planner such as Fast-Downward~\cite{helmert2006fast} alongside micro-level domain and problem definitions.
We denote a micro-plan for the $i$-th macro-action $A_i \in \mathcal{P}_{M}$ as $\tilde{\mathcal{P}_{\mu}}(A_i) = \{a_{i_1}, ..., a_{i_m} \}$, where $a_i$ denotes a micro-plan step (skills) and $m$ is the number of skills in the micro-plan necessary to ground $A_i$ in the  robot's world.
% $m = |A_i|$
Therefore, we can form a comprehensive micro-plan for the entire task $\mathcal{P}_{\mu}$ by combining the micro-plans for each macro-action in $\mathcal{P}_{M}$ as $\mathcal{P}_{\mu} = \{\tilde{\mathcal{P}_{\mu}}(A_1), \tilde{\mathcal{P}_{\mu}}(A_2), ..., \tilde{\mathcal{P}_{\mu}}(A_n) \}$, where $n = |\mathcal{P}_{M}|$.
It is important to note that a micro-plan for a given macro-PO depends on the configuration of the robot's environment.
To put it differently, although micro-POs are pre-defined, the permutation of these actions is solely derived from the linear planner, which distinguishes our hierarchical planning approach from others like HTNs, where methods must be pre-defined in addition to micro-POs (see Sec.~\ref{sec:exp-qual}).
}
\vspace{-0.1cm}
\section{Execution of a Manipulation Plan}
\label{sec:exe}
A manipulation plan is made up of a sequence of low-level actions that realizes the effects associated with high-level actions (functional units) in a FOON.
These low-level steps are automatically generated using the micro-level problem and domain definition, and they can be linked to motion primitives corresponding to skills.
In this work, we associate motion primitives with tuples known as \textit{action contexts}~\cite{agostini2020manipulation} that encode motion dependencies between consecutive actions in a plan for successful execution with an appropriate primitive.
\vspace{-0.25cm}
\subsection{Action Contexts}
\label{sec:action contexts}
% \todo{
% \textbf{Points to address}:
% \begin{itemize}
%     \item DMPs are tied to action contexts. Sometimes, collisions can occur, which can be remedied with re-planning or corrective strategies.
%     \item Re-planning would not change the high-level objective of FOON (in other words, not necessary for macro-planning, just micro-planning).
%     \item A better alternative would be to use motion planning to find trajectories.
% \end{itemize}
% }
%
An action context is a data structure that is used to associate a motion trajectory to a sequence of low-level skills. 
An action context $ac$ is represented as a tuple in the form of $ac = (a_{prev}, a_{now}, a_{next}, \mathcal{T})$, where $a_{now}$ refers to an action being executed, $a_{prev}$ and $a_{next}$ refer to preceding and proceeding actions, and $\mathcal{T}$ corresponds to its associated motion trajectory.
Each action ($a_{prev}, a_{now}$, or $a_{next}$) is made up of the PO name and its object arguments (as found by the planner), and a set or library of action contexts is denoted as $\mathcal{AC}$.
As in prior work~\cite{agostini2020manipulation}, trajectories are represented as dynamic movement primitives (DMPs)~\cite{ijspeert2013dynamical}, which use weights as forcing terms to preserve the shape of the original trajectory while allowing different initial and end positions of the robot's gripper.
\vspace{-0.3cm}
\subsection{Learning and Executing Action Contexts}
\label{sec:exe-ac}
\begin{figure}[t]
    \centering
    \includegraphics[width=5.5cm]{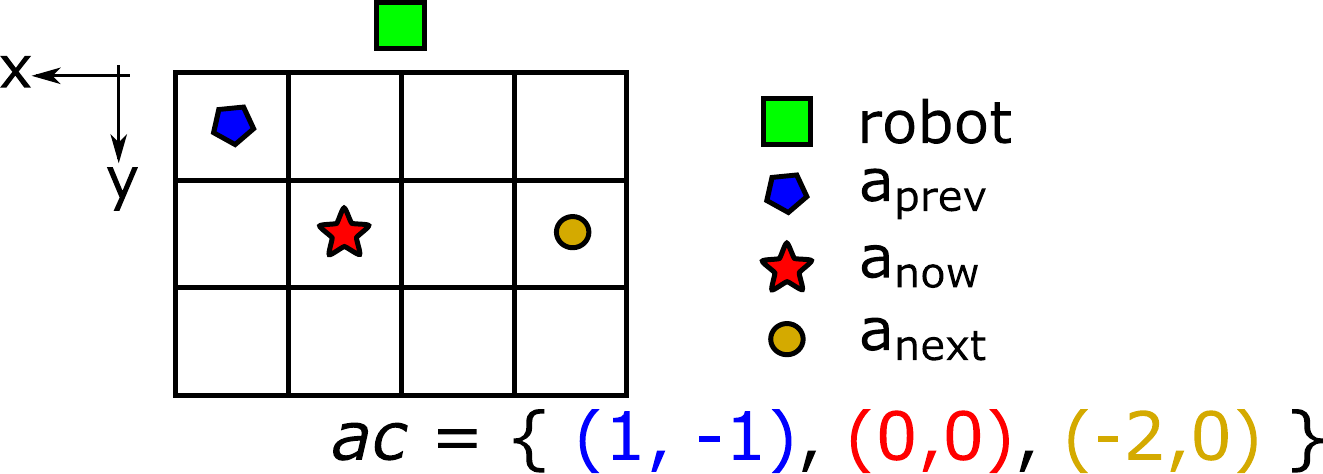}
    \caption{Illustration of action context generalization. An action context $ac$ can be generalized based on relative positioning of manipulations, where the location of $a_{now}$ is set as the origin $(0, 0)$. 
    % This allows us to reuse motion primitives assigned to each $ac$.
    The legend (on the right) indicates the symbols used to refer to the target location of each action in $ac$.\vspace{-0.4cm}
    }
    \label{fig:coord}
\end{figure}

When executing a micro-plan with $n$ actions (i.e., $\mathcal{P}_{\mu} = \{a_1, a_2, ..., a_n\}$) to achieve the effects of a macro-action $A \in \mathcal{P}_{M}$, a robot can search its library $\mathcal{AC}$ to derive the appropriate primitive for action $a_{t}$, given that a robot has executed a prior action $a_{t-1}$ and that it will then execute another action $a_{t+1}$ (if available).
% We denote the current action context as $ac$, where $ac = \{a_{prev}, a_{now}, a_{next}, g\}$, where $m$ is the action grounding mechanisms associated to the action context for the execution of $a_{now}$. 
% the primitive the robot needs to execute at time $t$.
% In other words, this would be equivalent to $ac^* = \{a_{t-1}, a_{t}, a_{t+1}, m^*\}$.
%
To select the appropriate DMP parameters $\mathcal{T}$, we first search for $ac \in \mathcal{AC}$ that matches the present context at a time-step $t$, where $a_{prev}$ is equal to $a_{t-1}$, $a_{now}$ is equal to $a_{t}$, and $a_{next}$ is equal to $a_{t+1}$. 
%
% However, it is likely that an exact action context cannot be found due to the unavailability of matches in $AC$ and the impracticality of demonstrating all possible $ac$ to the robot.
Action contexts are typically created from grounded actions observed in plan segments~\cite{agostini2020manipulation}.
However, encoding action contexts in this manner does not allow reuse in cases where the same motion dependencies are needed for a similar (but not equal) set of objects.  
% we transform each original (grounded) action context into an approximate one $\tilde{ac}$, and store it for future use. To elaborate, 
% Therefore, 
% In the likely case that an exact context match has not been found, we can use an approximate action context $\tilde{ac}$ in $\mathcal{AC}$. To elaborate, 
Hence, we generalize each $ac$ as a relative coordinate-like tuple, where $a_{t}$ is treated as the origin point (target), while $a_{t-1}$ and $a_{t+1}$ are treated as points relative to the origin (example shown as Fig.~\ref{fig:coord}). 
% find a \textit{similar} action context in terms of similar objects, actions, and relative positioning of manipulations.
% each action context $ac$ in $\mathcal{AC}$ % as well as $ac^*$ is generalized into 
This is inspired by previous work where planning operators were generalized using relative positions to targets in a grid world~\cite{agostini2017efficient}. 
% We provide an example in Fig.~\ref{fig:coord} to show how to derive these relative coordinates.
%
In addition, we define a mapping of objects to categories (viz., small, large, or wide) to define similarity across object types.
For instance, an $ac$ for a \textit{black pepper shaker} can apply to an object of similar size  like a \textit{salt shaker}.
With these concepts, we can extend action contexts to novel situations and reuse a suitable set of motion parameters $\mathcal{T}$.
However, if no $ac$ in $\mathcal{AC}$ matches the current micro-plan segment, a human demonstration is requested to create a new $ac$ and generate an associated set of DMP parameters~\cite{kulvicius2011joining}. 
As learning proceeds, the number of demonstrations decreases to zero and the robot eventually becomes fully autonomous \cite{agostini2020manipulation}.
%
% \subsection{Task Execution}
% As discussed before, a macro-level problem, which is closer to the abstract knowledge and concepts found in FOON, is broken down into a micro-level problem, which is concerned with deriving a sequence of manipulation actions that a robot can execute to solve the higher-level problem.
% Manipulation plans are executed using action contexts that maps segments of manipulation plans to motion parameters for trajectory planning (Sec. \ref{sec:action contexts})
% Manipulation actions are defined  can be learned and represented as motion primitives; in this work, we use dynamic movement primitives (DMP)~\cite{ijspeert2013dynamical} to learn and represent trajectories for a variety of skills. Based on a start and end position, a DMP can be used to approximate the original demonstrated trajectory; forcing terms are included that will retain details of the trajectory, where the more forcing terms (as weights) are used, the closer the learned trajectory will be to the original demonstration. We consider DMPs for skills such as picking, placing, pouring, cutting (in different ways -- e.g., chopping, slicing, and dicing), and mixing.

\vspace{-0.2cm}
\section{Evaluation}
\label{sec:exp}
%
% \subsection{Experimental Setup}
%% David: this will be for a follow-up work using multiple recipes, perhaps relying on object similarity:
% To demonstrate our framework of translating a FOON task tree or graph into a hierarchical manipulation planning problem, we selected three (3) simple cooking scenarios, whose graph annotations were originally from the FOON dataset.
% These scenarios were for the following recipes: 1) Greek salad, 2) miso soup, and 3) a Bloody Mary cocktail.
% To make it easier to replicate these recipes, while preserving the focus of demonstration and realism of the recipes, the graphs were simplified in some way, either by simplifying certain steps (e.g., some items were treated as being pre-processed) or reducing the number of used ingredients.
% Visualizations of these recipe graphs as well as demonstration videos of these scenarios are provided as supplementary material\footnote{TODO -- supplementary materials}.
%
% of translating a FOON task tree or graph into a hierarchical manipulation planning problem, 
To validate our approach, we perform cooking tasks via simulation in CoppeliaSim~\cite{coppeliaSim}.
% For this work, we selected the \textit{Bloody Mary cocktail} recipe from the FOON dataset.
% \rII{[C2.3]}~\rVI{[C6.11]
For this work, we created a universal FOON made of three subgraphs from the FOON dataset, from which we will perform hierarchical planning to prepare a \textit{Bloody Mary cocktail} and a \textit{Greek salad}.
The aim of macro-level planning is to extract a FOON-based plan for each goal (equivalent to a \textit{task tree}),
while that of micro-level planning is to find a task-level plan of robot-executable skills tailored to the state of the environment (viz. object locations and configurations).
% Although the knowledge in this graph is fixed, the manipulation plan obtained from macro- and micro-level planning can change based on the state of the environment (viz. object locations and configurations).
We thus show how this can be applied to randomly generated configurations of the scene while reliably and flexibly using action contexts and motion primitives.

% \begin{figure}[t]
%     % \vspace{4px}
%     \centering
%     \subfloat[Bloody Mary cocktail scenario]{
%     \includegraphics[width=5cm]{fig_config_cocktail-2.pdf}
%     \label{fig:config-A}}
    
%     % \hspace{1.5cm}
%     \subfloat[Greek salad scenario]{
%     \includegraphics[width=5cm]{fig_config_salad-2.pdf}
%     \label{fig:config-B}}
%     \caption{
%     \changes{
%     Layouts for the cocktail and salad tasks in CoppeliaSim.
%     Objects are as follows: 1) KUKA robot arm with Robotiq 2F-85 gripper, 2) a bottle of vodka, 3) a small table cell, 4) a long table cell, 5) a cup of Worcestershire sauce, 6) a black pepper shaker, 7) a cup of lemon juice, 8) a drinking glass, 9) a celery stick, 10) a knife, 11) a spoon, 12) a salt shaker, 13) a can of tomato juice, 14) a cup of ice, 15) a wide table cell, 16) a bottle of olive oil, 17) a cutting board, 18) a mixing bowl, 19) a cucumber, 20) a tomato, 21) a bowl of feta cheese, and 22) a bowl of olives.
%     }
%     }
%     \label{fig:sim_layout}
% \end{figure}
We evaluate our approach with a series of experiments to show that: 1) action contexts can be reused in novel scenarios, 2) FOON-based planning flexibly acquires plans for low-level situations that may not fully match that of the schema proposed by a FOON, and 3) FOON-based planning significantly improves computation time over classical and HTN planning.
To address 1) and 2), we measure the average success rate of plan execution for randomized scenes and ingredient subsets, while to address 3), we measure computation time as the overall time taken by the planners to find a solution.
An illustration of the universal FOON and demonstration videos for Sec.~\ref{sec:exp-qual} and {\ref{sec:exp-transfer} are provided as supplementary materials.\footnote{Supplementary Materials -- \url{https://davidpaulius.github.io/foon-lhpe}}

% \AG{Showing the computation time with and without FOON is good to evaluate scalability but it would not suffice as a thorough experimental evaluation. Is there any difference regarding the rate of success with and without FOON? Other possible evaluations: show how the framework is able to prepare cocktails under different physical configurations (transferability). E.g. define XX randomly generated initial physical configurations and calculate the rate of success. Also, evaluate the capability of the approach to generate different FOON-plans. Computation time to find a FOON-plan. Rate of (execution) success for different FOON-problems. It would be also informative to describe how many demonstrations were needed for the system to become fully autonomous, how many action contexts, etc.}
% \vspace{-1cm}

\subsection{Experimental Setup}
\label{sec:exp-setup}

Using CoppeliaSim, we designed simple table-top environments with objects and utensils that will be manipulated by a single KUKA LBR iiwa 7 R800 robot arm equipped with a Robotiq 2F-85 gripper.
% \changes
Fig.~\ref{fig:sim_layout} shows the layout for the cocktail and salad tasks.
To make it easier to replicate the recipes while preserving realism, we simplified certain steps in the recipe's FOON for one-armed manipulation; for example, rather than squeezing a lemon for juice, we provide a cup of lemon juice in the scene.
This is similar to the cooking principle of \textit{mise en place}.
We also fashioned objects, such as the cutting board or knife, for robotic manipulation.
% We also assume that there is only a single instance of each object type.
\changes{We make several assumptions for perception as mentioned in Sec.~\ref{sec:tamp}.
First, objects are placed on cells that discretize the surface; since there are objects of varying sizes (i.e., small, long, and wide), we designed appropriately sized table cells upon which they may be placed.
Second, we assume that ingredients in cups or shakers are initialized and symbolically propagated from their starting containers (e.g., \textit{ice} is in the \textit{cup of ice}).}
% Non-liquid ingredients such as \textit{olives} are detectable.
% (e.g., (3), (4) and (15) in Fig.~\ref{fig:sim_layout}).
% Furthermore, due to the robot's limited reach to objects in table cells closest to it (highlighted in yellow, example labeled as ()
% In the scene, there are a total of 21 cells, where large table cells are used for placing large objects (marked as (6) to (8) in Fig.~\ref{fig:sim_layout}).
We use the following off-the-shelf planners in our experiments: Fast-Downward~\cite{helmert2006fast}
and $\text{PANDA}_{Pi}$\footnote{$\text{PANDA}_{Pi}$ Planner -- \url{https://panda-planner-dev.github.io/}} (which uses HDDL~\cite{holler2020hddl}).

\begin{figure}[t]
    \centering    
    \includegraphics[width=0.9\columnwidth]{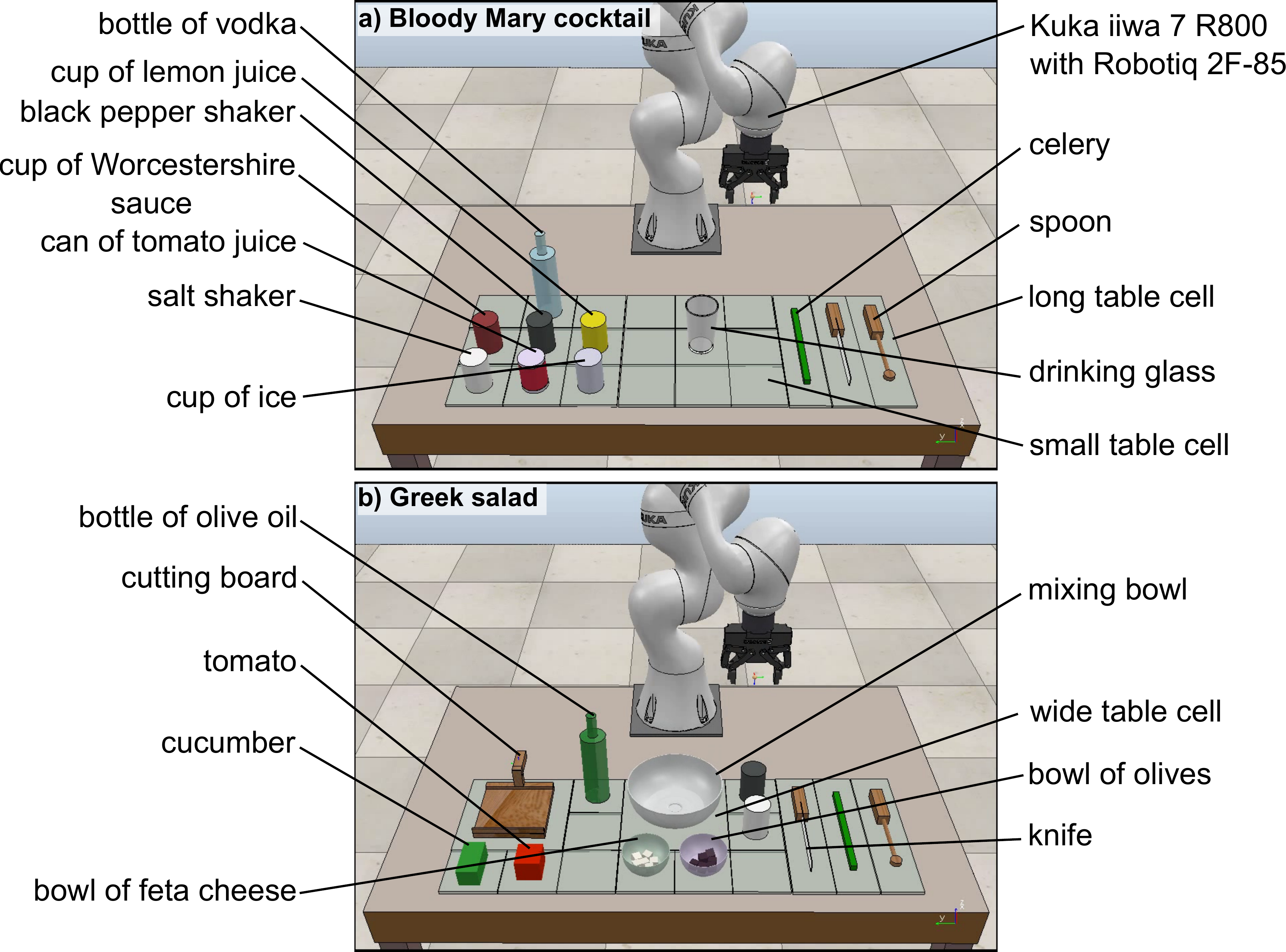}
    \caption{
        Layouts for the cocktail and salad tasks in CoppeliaSim.    
    }
    \label{fig:sim_layout}
\end{figure}

\begin{figure*}
    \centering
    \includegraphics[width=0.97\textwidth]{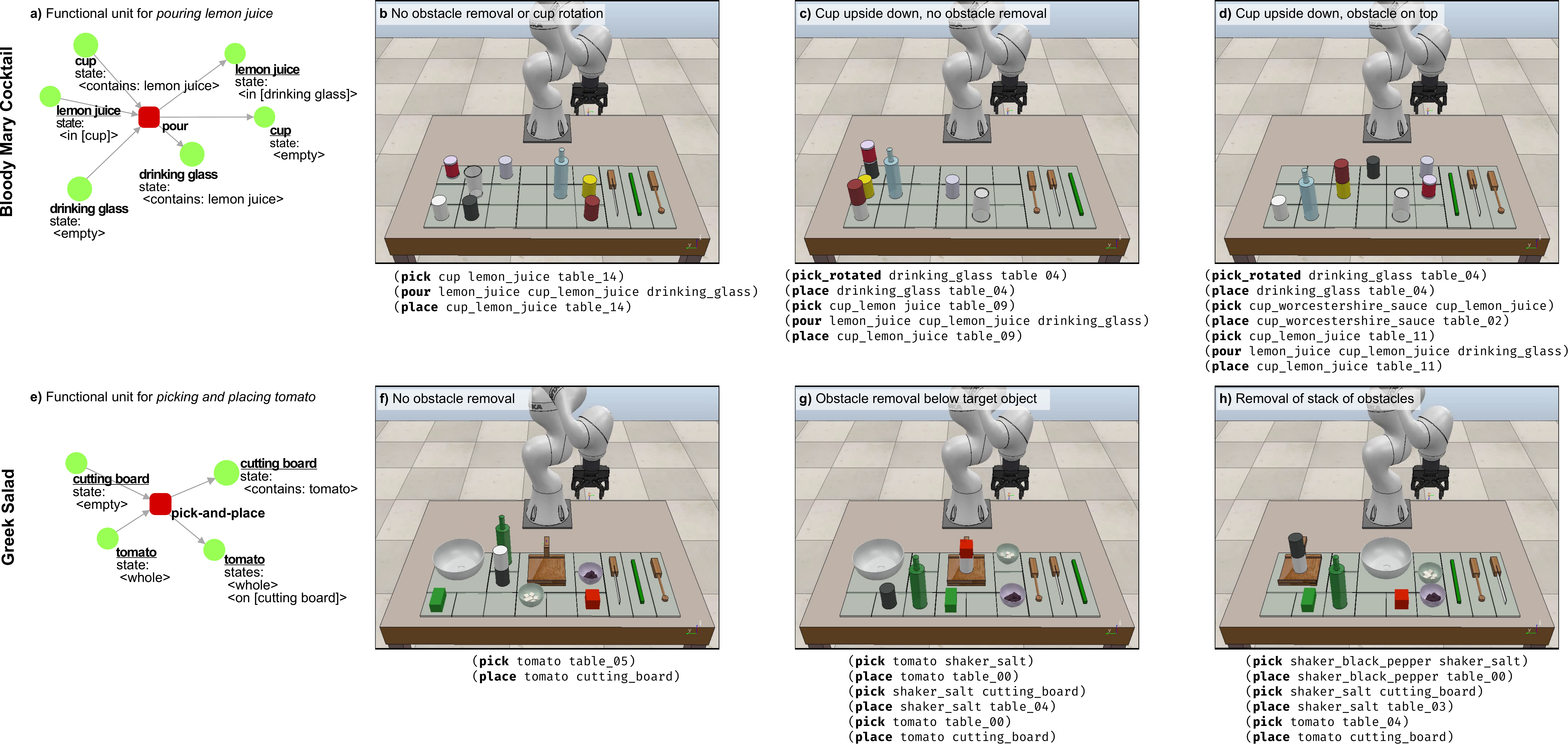}
    \caption{Qualitative examples of plan variations for cocktail and salad recipes.
    We highlight various micro-plans for the same macro-level objective (i.e., functional unit).
    For the cocktail scene, the \textit{macro-PO} is {\scriptsize{\tt{(pour\_lemon\_juice)}}} to pour lemon juice, while for the salad scene, the \textit{macro-PO} is {\scriptsize{\tt{(pick\_and\_place\_tomato)}}} for putting a tomato on the cutting board.
    Videos for each qualitative example are provided in supplementary materials.
    \vspace{-0.4cm}
    }
    \label{fig:qual-combined}
\end{figure*}

\vspace{-0.3cm}
\subsection{Plan Generation for Variable Object Configurations}
\label{sec:exp-qual}
First, we demonstrate how our approach finds micro-plans for varying object configurations and constraints for the same macro-PO.
We perform these experiments on both cocktail and salad scenarios. 
Fig.~\ref{fig:qual-combined} shows various scene configurations and micro-plans for \pred{(pour\_lemon\_juice)} in the cocktail scene (Figs.~\ref{fig:qual-combined}b--d) and \pred{(pick\_and\_place\_tomato)} in the salad scene (Figs.~\ref{fig:qual-combined}f--h) along with their respective micro-plans. 
These macro-POs are equivalent to the functional units shown as Figs.~\ref{fig:qual-combined}a and \ref{fig:qual-combined}e (ignoring irrelevant ingredients).

The cases for the \textit{pour} task are as follows: 1) the objects are clear for pouring (Fig.~\ref{fig:qual-combined}b); 2) the drinking glass requires rotation before pouring (Fig.~\ref{fig:qual-combined}c); and 3) the drinking glass requires rotation and the ingredient (cup of lemon juice) is blocked (Fig.~\ref{fig:qual-combined}d).
% ; (another alternative, 4) the glass is rotated and blocked, and the ingredient is also blocked).
The cases for the \textit{pick-and-place} task are as follows: 1) the cutting board is free of obstacles for placing the tomato on top of it (Fig.~\ref{fig:qual-combined}f); 2) the tomato is obstructed by a salt shaker between it and the cutting board (Fig.~\ref{fig:qual-combined}g); and 3) the cutting board has a stack of obstacles on it that need to be removed prior to placing the tomato (Fig.~\ref{fig:qual-combined}h).

% \changes
% {
Fig.~\ref{fig:qual-combined} shows generated plans for different states, yet the \textit{same} macro-level objective.
We provide links to videos for each micro-plan execution in our supplementary materials.
\changes{When defining this domain as an HTN, we must define methods for every possible way of executing macro-actions. 
We observed this when implementing an HTN using $\text{PANDA}_{Pi}$ and HDDL: just for pouring, we had to account for 8 variations of sub-task sequences, such as when an obstacle was present (blocking a container) or if the target container was not properly oriented. Further, we must define a set of (primitive) actions for an HTN; we use the same set of micro-POs to define these actions.
In addition to these actions, we had to define a total of 19 methods for HTN planning.
Our approach instead relies solely on a linear planner to find a geometrically feasible micro-plan, whose order is determined by the planner.}
% }
% }

% Salad scenario
% \AG{TODO:ONLY IF THERE IS TIME FOR THIS: For the salad scenario: 1) cutting board ready for cutting; 2) cutting board blocked with a pile of two objects; 3) WHAT ELSE?}

% \vspace{-0.1cm}
\subsection{Transferability to New Scenarios}
\label{sec:exp-transfer}
%\AG{We are not only showing transferability of ACs but also that, having POs, we are able to define different sequences of actions to solve different physical constraints for planning and execution. This cannot be done using only FOON.}
To demonstrate transferability, we perform two kinds of experiments over 25 trials in variable scenarios: 1) \textit{whole recipe} execution, using all ingredients in the original recipe; and 2) \textit{partial recipe} execution, using random ingredient subsets.
Although the \textit{same} object-level plan is found across all trials, each trial results in different manipulation plans due to the configuration of objects (Fig.~\ref{fig:sim_layout}) in the scene (cases such as those in Sec.~\ref{sec:exp-qual}). 
In addition, we show that FOON can be flexibly modified at the PDDL level to plan for novel scenarios using fewer objects without creating a new FOON via partial recipe execution.
% we expect a different manipulation plan across trials, as the kitchen layout will be reconfigured such that will be placed in random locations and orientations .
% As a result, manipulation plans can range between 26 and 28 micro-level steps.
% For each round, 25 trials of planning and execution are conducted via simulation.
A trial is successful if all objects are manipulated with a suitable action context and motion primitive while avoiding collisions that may cause remaining steps to fail.
For example, if the robot knocks a bottle out of the workspace (i.e., table cells) before pouring, then the robot is unable to complete its corresponding macro-PO.
Objects stacked on top of others would be placed in a free spot after use to avoid further removing them for remaining steps.

Since objects are randomly configured at the start of each trial, the robot has to rely on learned action contexts.
We collected a total of 703 action contexts from demonstration (635 from the cocktail task and an additional 68 from the salad task), which can be generalized using the method from Sec.~\ref{sec:exe-ac}.
We summarize our results in Table~\ref{tab:exp}.
% \changes
In the cocktail task, robot execution was 96\% successful for whole execution and 92\% for partial execution; in the salad task, robot execution was 80\% successful for whole execution and 84\% for partial execution.
In some trials, the robot failed to complete the task due to object collision, as trajectories encoded by action contexts are not adapted to avoid collisions with objects lying in between manipulated ones.
% (as in).
% This is especially prevalent in the salad task, which is a longer-horizon task with an average plan length of 35 steps.
Despite the lack of collision avoidance mechanisms~\cite{urbaniak2020combining}, however, the stored shapes were enough to avoid collisions in most cases. % without the need to replan.
\begin{table}[t]
    \centering
    {
    \scriptsize
    \renewcommand{\arraystretch}{1.1}
    \caption{
    % \AG{Redo Fig. 9 with 25 randomized configurations and find the average planning times (with and without FOON/macro-level planning).}
    Success rates with randomized configurations of scene objects for whole and partial recipe execution}
    \begin{tabular}{cM{1.1cm}M{1.3cm}M{1.6cm}M{0.9cm}}
    \toprule[1pt]
    \textbf{Task} & \textbf{Execution Type} & \textbf{Avg. Plan Length} & \textbf{No. Successful Trials} & \textbf{\% Success}  \\
    \midrule
    \multirow{2}{*}{\textit{Cocktail}} & Whole & $27.9 \pm 1.35$ & $24/25$ & $96\%$  \\
    & Partial & $19.8 \pm 3.57$ & $23/25$ & $92\%$  \\
    \midrule
    \multirow{2}{*}{\textit{Salad}} & Whole & $34.6 \pm 1.78$ & $20/25$ & $80\%$  \\
    & Partial & $24.9 \pm 5.33$ & $21/25$ & $84\%$  \\
    \bottomrule[1pt]
    \end{tabular}
    \label{tab:exp}
    }
\end{table}

\subsection{Comparison to other Planning Methods}

% \rIX{[C-9.4]
An advantage of using functional units to define PDDL problems is that it simplifies planning, where, rather than composing a single problem definition, our approach transforms each functional unit into smaller problem definitions, which benefits in a significantly reduced time complexity.
To support this claim, we compared the average computation time over 10 cocktail scenes for three flavours of planning: (1) FOON-based planning, where we transform each functional unit into macro-problems (our approach in this work); (2) classical planning, where a single problem file is defined with goals of $n$ functional units (where $n$ ranges from 1 to $|\mathcal{P}_{M}|$); and (3) HTN planning implemented with $\text{PANDA}_{Pi}$ and HDDL~\cite{holler2020hddl}, where we define an HTN with tasks and methods using micro-PO definitions.
% }
% \todo{Can we quantify the effort put for handcrafting the HTN?}
We use A$^*$ search with two heuristics: landmark cut (LMCUT) and Fast Forward (FF)\footnote{Details on heuristics -- {\url{https://www.fast-downward.org/Doc/Evaluator}}}.
% LMCUT is an admissible heuristic that finds optimal plans, while FF is non-admissible yet it can be used to find acceptable plans.
Running times were measured on a machine running Ubuntu 20.04 with 16 GBs of RAM and an Intel Core i5-8300H processor.
% \changes
{
A maximum allotted time of 20 minutes was set for each trial.
We plot our findings as Fig.~\ref{fig:timings} using a logarithmic scale to highlight the difference in time complexity between the three approaches.
%
%
% \AG{This experiment needs to be re-done. Define 10 or more (ideally 25) initial situations where the goals are defined by the first, second, third, and so on, functional unit. With-FOON: this case needs a perception stage after executing each functional unit to define the ini state for micro-planning. Without-FOON: just run it linearly with goals given by each functional unit (no perception after each functional unit execution).}

\begin{figure}[t]
    % \vspace{2px}
    \centering
    \includegraphics[width=0.97\columnwidth]{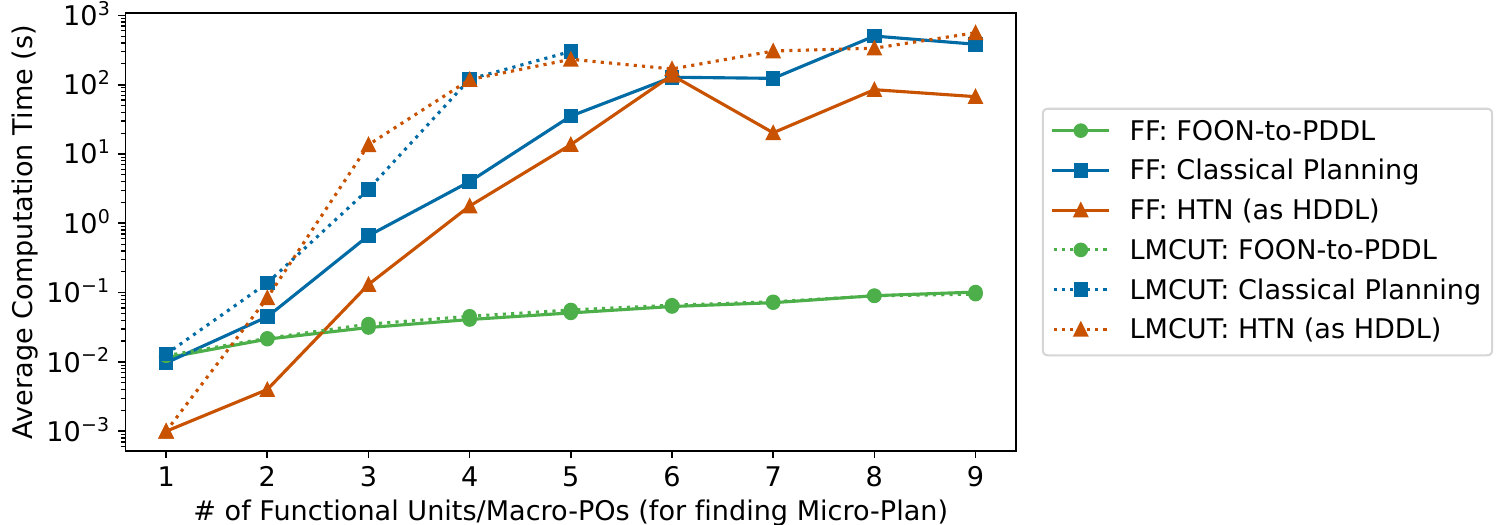}
    \caption{Graph showing average planning times for solutions found over 10 cocktail scenes for 1) FOON-based planning, 2) classical planning, and 3) HTN planning (HDDL). We compare using landmark cut (LMCUT) and Fast Forward (FF) heuristics for A$^*$ search.
    This graph uses a log-scale to highlight timing differences.
    Plans beyond 5 functional units were not found within the allotted time using classical planning and LMCUT.
    }
    \label{fig:timings}
\end{figure}

\begin{figure}[t]
    % \vspace{2px}
    \centering
    \includegraphics[width=0.97\columnwidth]{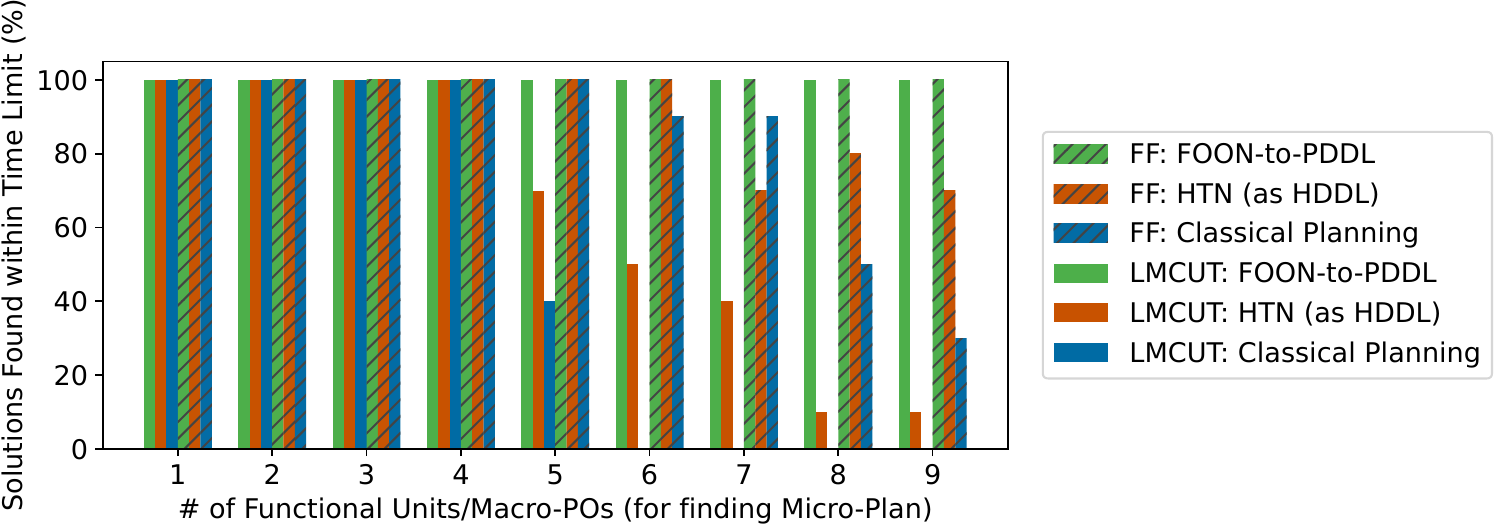}
    \caption{Graph showing the percentage of scenes for which a solution was found within the allotted time of 20 minutes (best viewed in colour).
    \vspace{-0.4cm}
    }
    \label{fig:completed}
\end{figure}

% \rIX{[C-9.4]
We can observe that FOON-based planning finds plans in significantly less time than the other methods, as the planner operates with smaller, independent search spaces, which is facilitated by perception for state updates.
HTN planning found plans in shorter time for smaller problem sizes, but computation times gradually increase for problems larger than 3 functional units.
It is also important to note that Fig.~\ref{fig:timings} shows averages for solutions that were found within 20 minutes; however, many problems of sizes beyond 4 functional units could not be found within the allotted time (see Fig.~\ref{fig:completed}).
% Using LMCUT for classical planning took a significantly longer time to find a solution, so much that plans were not found for problems larger than 5 functional units.
% In addition, despite non-admissibility of FF, classical planning could not perform as well as FOON-based planning.
% }
}

Our approach exploits optimal heuristics on smaller problem sets, which allows a robot to find and execute a plan in real-time.
Further, perception can be used between macro- and micro-actions to monitor the state of the environment.
Finally, FOON schematically enforces a high-level ordering of actions.
One key example that requires such ordering is \textit{mixing}.
At the macro-level, mixing requires ingredients in a container, but at the micro-level, the only requirement is that the container is free of obstacles on top of it, as it results in the \textit{container} being mixed (\pred{is-mixed \var{cnt}}) rather than its contents being mixed.
Hence, without a macro-plan, we may acquire a plan where mixing is done before adding all ingredients.

\vspace{-0.2cm}
\section{Conclusion}
\label{sec:con}
% \todo{
% \textbf{Points to address}:
% \begin{itemize}
%     \item It is important to note that we are not proposing a new task planning approach, rather mechanisms to ground the easy-to-obtain and rich task representation of FOONs for service robot executions exploiting the efficiency of off-the-shelf linear planners and the advantages of object-centered predicates representation for the generation of feasible plans.
%     % We also want to motivate the use of plan schemas for bootstrapping PDDL planning.
% \end{itemize}
}
% \todo{We are not proposing a new task planning approach, rather mechanisms to ground the easy-to-obtain and rich task representation of FOONs for service robot executions exploiting the efficiency of off-the-shelf linear planners and the advantages of object-centered predicates representation for the generation of feasible plans.}

% \changes{We introduce an approach to ground the easy-to-obtain, rich knowledge representation of \textit{functional object-oriented network} (FOON) for robotic execution of long-horizon tasks in variable scenarios. This approach exploits the efficiency of off-the-shelf linear planners and the advantages of object-centered predicates representation and action contexts for the generation of feasible plans.}

\changes{In summary, we introduce an approach to ground rich, object-level knowledge in \textit{functional object-oriented networks} (FOON) for robotic execution of long-horizon tasks in variable scenarios. 
This approach exploits the efficiency of off-the-shelf planners, action contexts, and object-centered representation in PDDL for the generation of geometrically feasible plans.}
% We introduce an approach to combine domain knowledge from the functional object-oriented network (FOON) and classical planning via PDDL to perform task planning for robotic execution.
This is done by a two-step hierarchical decomposition that deconstructs a FOON's functional units into planning operators and predicates in PDDL notation, allowing us to leverage off-the-shelf planners and existing search algorithms.
Using object-level representations like FOON to bootstrap task and motion planning allows us to quickly generate flexible solutions that are tailored to the state of the robot's environment.
% This is because the knowledge in FOON is domain-independent enough to be extended to new object types, environments, agents, or robots, and it can be made domain-specific to suit a specific set of robots, skill repertoires, and object instances.
% , which is not necessary to be present in an object-level representation such as FOON.
%

% \rX{[C-10.2]
% Representations closer to human language and understanding such as FOON can be defined by us humans in an abstract yet intuitive way for robotics.
% Object-level representations like FOON are designed without committing to a specific robotic implementation of task execution.
% In other words, knowledge in FOON is domain-independent enough to be extended to new object types, environments, agents, or robots, and it can be made domain-specific to suit a specific set of robots, skill repertoires, and object instances.
% Rather, domain knowledge becomes domain- and robot-specific after translation from a macro-level to a micro-level description.
% }

\vspace{0.3cm}

\subsection{Limitations and Future Work}
Despite the exceptional performance of our approach at long-horizon task planning and execution, there are several limitations that we plan to address as future work.
\changes{
One issue is the open-loop nature of the robotic execution, which is unsuited to handle unexpected contingencies inherent to real-robot scenarios when executing micro-plans, such as collisions with objects or external changes to the environment.
We will explore re-planning options in the same vein of prior work~\cite{agostini2017efficient} and include geometric feedback via motion planning in real-world settings for micro-plan execution.} 
% Additionally, further demonstration of our framework will be conducted with real robots so as to validate the robustness of our system when operating under the uncertainties inherent to real world settings. 
% \rVI{[C6.2]}~\rVI
% \changes
Although the DMPs tied to action contexts can reproduce the shape and orientation of trajectories for demonstrated actions, they do not guarantee collision-free executions. 
We plan to incorporate mechanisms to adapt motion primitives for obstacle avoidance~\cite{urbaniak2020combining}.
\bibliographystyle{IEEEtran}
\bibliography{ref}

\end{document}